\documentclass[10pt,twocolumn,letterpaper]{article}

 \usepackage{cvpr} 
\usepackage{subcaption}

\usepackage{array}
\usepackage[table]{xcolor}
\newcolumntype{H}{>{\setbox0=\hbox\bgroup}c<{\egroup}@{}}

\definecolor{cvprblue}{rgb}{0.21,0.49,0.74}
\usepackage[pagebackref,breaklinks,colorlinks,allcolors=cvprblue]{hyperref}
\usepackage{algpseudocode}
\usepackage{algorithm}
\def\paperID{-} 
\def\confName{CVPR}
\def\confYear{2026}
\newcommand{\ours}[1][]{Sparse-LaViDa}

\def\etc{\emph{etc}}

\title{\ours: Sparse  Multimodal Discrete Diffusion Language Models}

\author{Shufan Li$^{1,2,*}$, Jiuxiang Gu$^{1}$, Kangning Liu$^{1}$, Zhe Lin$^{1}$ \\ Zijun Wei$^{1}$, Aditya Grover$^{2}$, Jason Kuen$^{1}$ \\
$^1$Adobe~~$^2$UCLA \\
}

\begin{document}
\maketitle
\begin{abstract}
Masked Discrete Diffusion Models (MDMs) have achieved strong performance across a wide range of multimodal tasks, including image understanding, generation, and editing. However, their inference speed remains suboptimal due to the need to repeatedly process redundant masked tokens at every sampling step. In this work, we propose \ours, a novel modeling framework that dynamically truncates unnecessary masked tokens at each inference step to accelerate MDM sampling. To preserve generation quality, we introduce specialized register tokens that serve as 
compact representations for the truncated tokens. Furthermore, to ensure consistency between training and inference, we design a specialized attention mask that faithfully matches the truncated sampling procedure during training. Built upon the state-of-the-art unified MDM LaViDa-O, \ours~achieves up to a 2$\times$ speedup across diverse tasks including text-to-image generation, image editing, and mathematical reasoning, while maintaining generation quality.
\end{abstract}    
\section{Introduction}
\label{sec:intro}

Improving visual understanding and generation capabilities has been a major focus of artificial intelligence research. A common paradigm is to employ autoregressive (AR)
vision-language models (VLMs)
for visual understanding tasks such as question answering, and diffusion models for visual generation tasks such as text-to-image generation and image editing. Recently, there has been rising interest in building unified multimodal models capable of both understanding and generation~\cite{zhou2024transfusion,deng2025emerging,wu2025vilau,ma2025unitok}. These models often surpass the performance of single-task systems because they enable understanding and generation capabilities to mutually benefit from shared representations under a unified framework, which is especially advantageous for tasks requiring both abilities, such as image editing.

\begin{figure}
    \centering
    \includegraphics[width=1.0\linewidth]{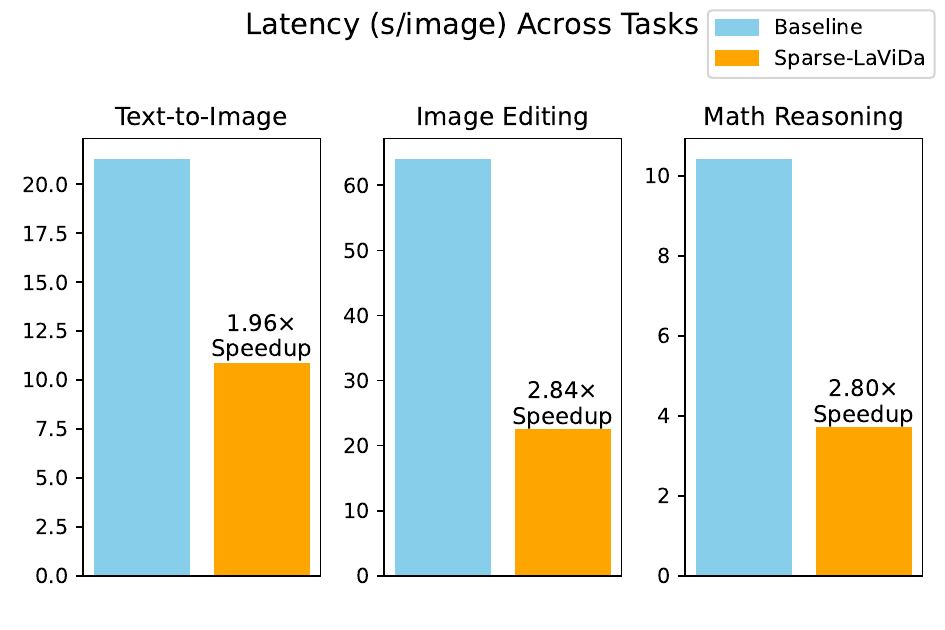}
    \caption{\textbf{We propose \ours}, a novel modeling technique for unified multimodal masked discrete diffusion models. \ours~achieves substantial speedup across a wide range of tasks, including text-to-image generation, image editing, and visual math reasoning, compared with the baseline LaViDa-O.}
    \label{fig:main_performance}
\end{figure}

\begin{figure*}[ht]
    \centering
    \includegraphics[width=1.0\linewidth]{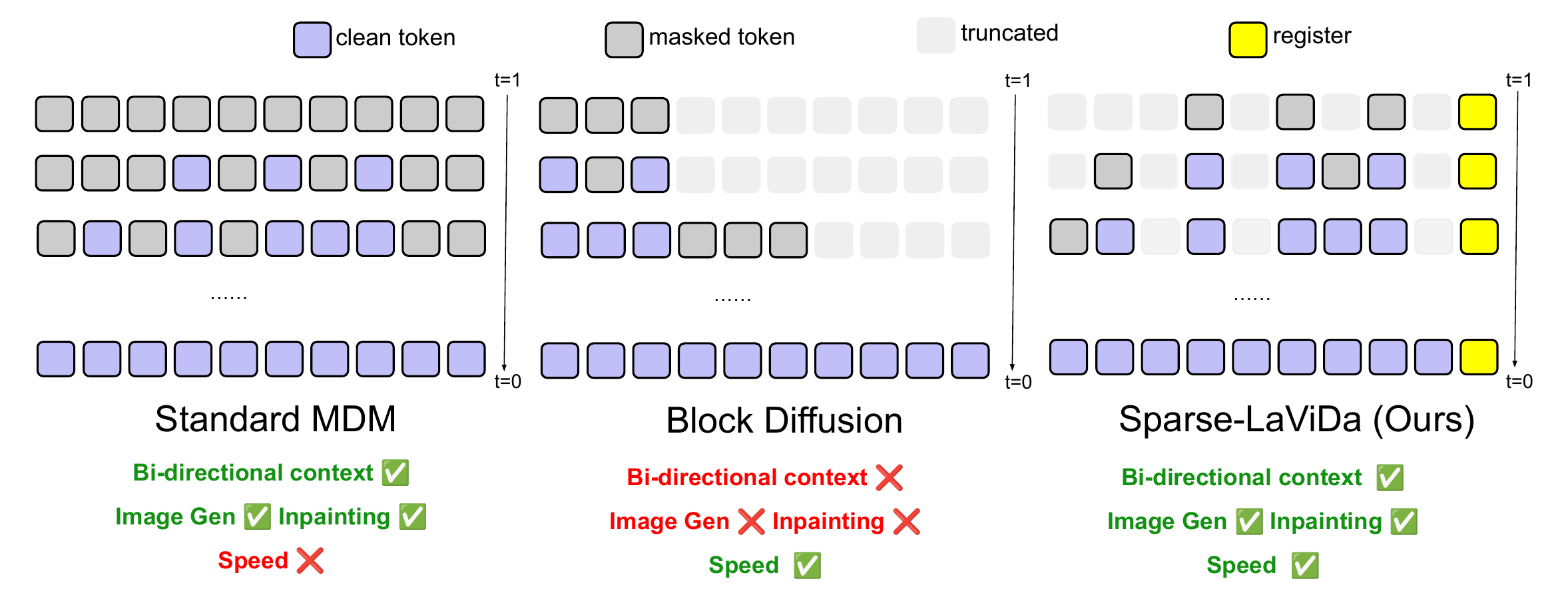}
    \caption{\textbf{Overall design of \ours.} \textit{Left:} Vanilla MDMs materialize all masked tokens and support arbitrary-order decoding (top-down). Unlike AR models, they have bidirectional context and naturally support tasks such as image generation and inpainting. \textit{Middle:} Block Diffusion truncates redundant masked tokens from the right but imposes a left-to-right generation order using a block-causal attention mask, losing many benefits of MDMs. \textit{Right:} \ours~is an alternative parameterization of vanilla MDMs. It preserves all benefits of standard MDMs while achieving the efficiency gains of Block Diffusion by allowing mask truncation at arbitrary positions. Special register tokens serve as compressed representations of truncated tokens.}
    \label{fig:teaser}
\end{figure*}

Early works such as Transfusion~\citep{zhou2024transfusion} and BAGEL~\citep{deng2025emerging} build unified multimodal models by combining 
AR
VLMs
with continuous diffusion models to handle understanding and generation tasks respectively. More recently, Masked Diffusion Models (MDMs) have emerged as a promising alternative, offering a unified modeling framework for both text and image generation~\cite{li2025lavida,yang2025mmada,li2025lavidao,yu2025dimple,shi2025muddit}. 
Concretely, MDMs represent both text and images as sequences of discrete tokens. Given such a sequence, the forward diffusion process gradually converts it into a fully masked sequence. An MDM learns the reverse process by predicting the distribution of original tokens at the masked positions. To sample from MDMs, we start with an all-mask sequence and iteratively unmask tokens to obtain a clean sequence. This formulation brings several advantages over
AR
models, such as faster inference via parallel decoding, controllable generation, and bidirectional context,~\etc. (Figure \ref{fig:teaser} Left)
Notably,
the unified MDM LaViDa-O~\citep{li2025lavidao} achieves strong performance across a wide range of image understanding and generation tasks.

Despite supporting parallel decoding, existing MDMs still face major efficiency limitations. First, they rely on full attention instead of the causal attention used by AR models. While this enables bidirectional context and naturally supports tasks such as text infilling and image inpainting, it prevents the use of KV-cache acceleration during inference. Second, they must repeatedly process the full sequence, including many redundant masked tokens, at every sampling step. For example, if an image is represented by 1024 tokens, the model process all 1024 tokens at each diffusion step, even though only a small subset is unmasked at a time.
Recently,
Block Diffusion~\citep{arriola2025block} was proposed to improve MDM efficiency by constraining parallel decoding to a left-to-right block-causal order (Figure~\ref{fig:teaser}, Middle). While effective for language modeling, this left-to-right generation scheme is poorly suited for image generation and editing tasks, where tokens do not follow a natural ordering. Furthermore, its block-causal attention mask eliminates bidirectional context,a key advantage of MDMs, making tasks such as inpainting difficult.

To address these limitations, we propose \ours, a novel modeling framework that improves the efficiency of MDMs by supporting KV-cache usage and enabling truncation of arbitrary subsets of redundant tokens during inference. Concretely, \ours~introduces three key innovations. First, we propose a sparse parameterization of MDMs that represents partially masked sequences without materializing all masked tokens. At each sampling step, \ours~takes only the prompt tokens, previously generated tokens, and a selected subset of masked tokens that need to be decoded, in contrast to vanilla MDMs which always materialize all masks. Second, we introduce special register tokens that serve as compressed representations of truncated tokens and help recover modeling capacity lost due to truncation. Finally, we design a step-causal attention mask that enables KV-cache support during inference while allowing efficient training. Unlike the block-causal mask used in Block Diffusion, our step-causal attention mask preserves the bidirectional context essential for image generation, editing, and inpainting. These components are illustrated in Figure~\ref{fig:teaser} 
(Right).

To validate the effectiveness of \ours, we conduct extensive experiments across image generation, understanding, and editing tasks. \ours~achieves significant efficiency gains, including a $1.96\times$ speedup on text-to-image generation, $2.80\times$ speedup on image editing, and $2.84\times$ speedup on visual math reasoning, while maintaining generation quality comparable to the unified MDM LaViDa-O~\cite{li2025lavidao} (Figure \ref{fig:main_performance}). Notably, these improvements are achieved on top of many optimizations already employed by LaViDa-O, such as token compression.

\section{Related Work}
\begin{figure*}[ht]
  \centering
  \begin{subcaptionbox}{Inference with Sparse Parameterization \label{fig:modality-aware-masking}}[0.64\textwidth]
    {\includegraphics[width=\linewidth]{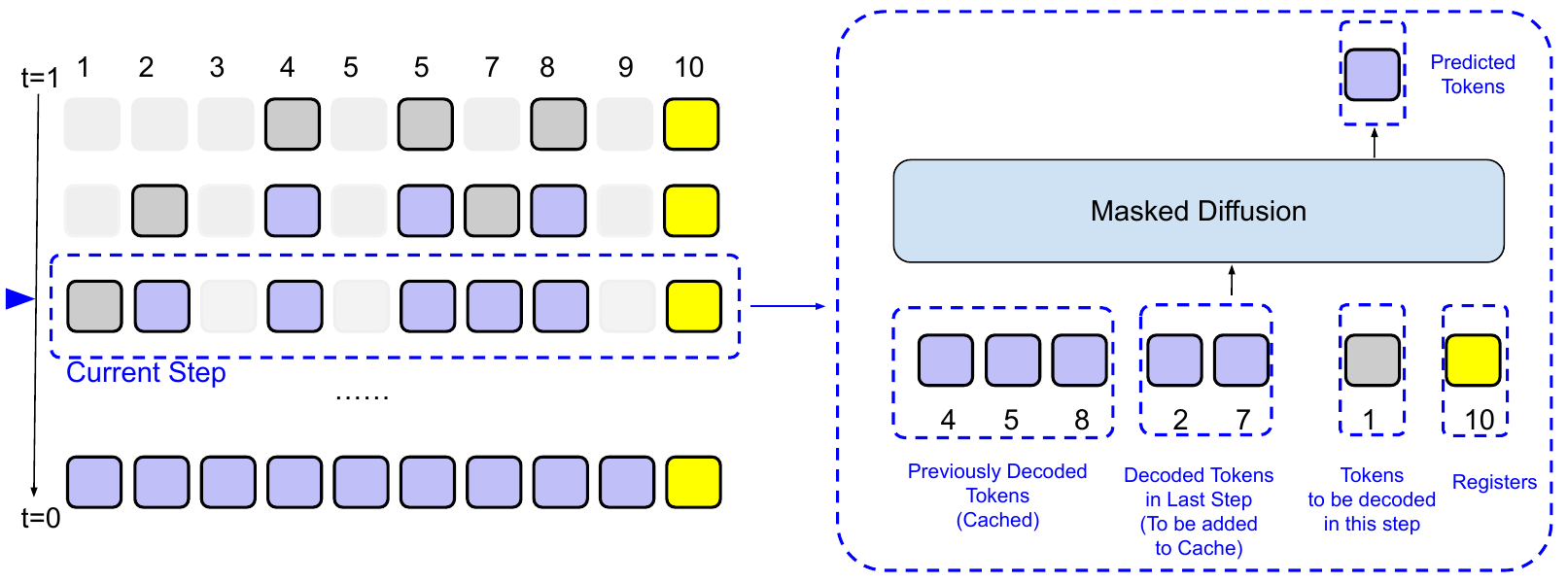}}
  \end{subcaptionbox}
  \begin{subcaptionbox}{Inference Attention Mask of \ours \label{fig:stratified-sampling}}[0.32\textwidth]
    {\includegraphics[width=\linewidth]{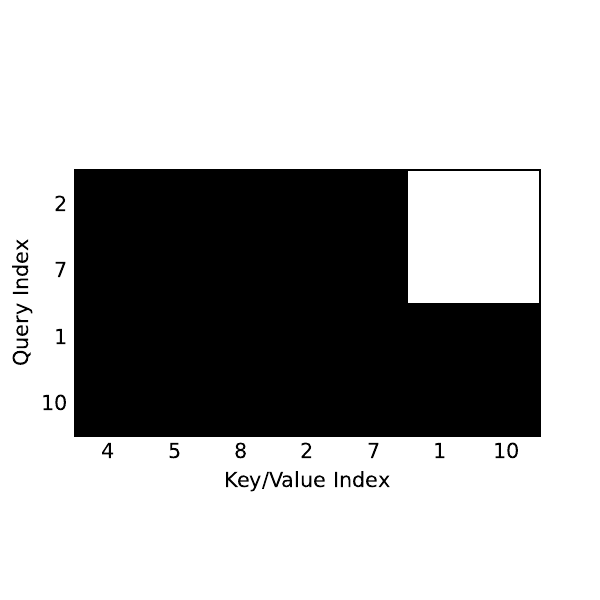}}
  \end{subcaptionbox}
  \caption{\textbf{Inference pipeline of \ours.} 
  (a) At a specific decoding step (top-down), the input of \ours~consists of four types of tokens: 
  (1) previously decoded tokens stored in the KV cache; 
  (2) newly decoded tokens from the previous step, which will be added to the cache at this step; 
  (3) masked tokens to be decoded at the current step; and 
  (4) register tokens. 
  (b) During sampling, we apply a specialized attention mask such that tokens of type~(2) cannot attend to tokens of type~(3) or~(4). 
  In this example, tokens at positions~2 and~7 are of type~(2) (newly decoded), while tokens at positions~1 and~10 correspond to types~(3) and~(4) (mask and register), respectively.}
  \label{fig:method_infer}
\end{figure*}

\subsection{Masked Diffusion Models}
\label{sec:mdm_related}

Early works on masked modeling, such as BERT~\citep{devlin2019bert} and MAE~\citep{he2022masked}, focus on learning semantically rich representations through masked autoencoding. MaskGIT~\citep{chang2022maskgit} and Muse~\citep{chang2023muse} were among the first to apply masked modeling to image generation, using VQGAN~\citep{esser2021taming} to convert images into sequences of discrete tokens. Despite some success, these early approaches lacked a strong theoretical foundation and relied heavily on heuristics. Recently,
MDMs~\citep{austin2021structured-d3pm,sahoo2024simple,lou2023discrete-sedd,shi2024simplified} have established a principled theoretical framework by formulating the masking and unmasking processes as forward and reverse discrete diffusion processes. This provides a unified and mathematically grounded approach for training and sampling in masked generative models. Building on this foundation, several works
such as Mercury~\citep{khanna2025mercury}, LLaDA~\cite{nie2025large} and Dream~\citep{dream2025} have successfully scaled MDMs to large-scale language modeling, achieving performance comparable to autoregressive counterparts while offering advantages such as bidirectional context and parallel decoding.  
Further extensions, including LaViDa, LaViDa-O, and MMaDa~\citep{li2025lavida,li2025lavidao,yu2025dimple,shi2025muddit,yang2025mmada}, have expanded MDMs to multimodal understanding and generation tasks, achieving impressive results.

Formally, given a sequence of discrete tokens $X_0 = [X_0^1, X_0^2, \dots, X_0^L]$, where $L$ denotes the sequence length, the forward masked diffusion process $q(X_t|X_s)$ progressively replaces clean tokens with masked tokens over the time interval $[0,1]$, with $1 \ge t \ge s \ge 0$. At $t = 1$, the sequence $X_1 = [M, M, \dots, M]$ consists entirely of masked tokens. For intermediate steps where $0 < t < 1$, $X_t$ contains a mixture of clean and masked tokens. A neural network $p_\theta$ is trained to model the reverse process $p(X_s|X_t)$. The masked diffusion objective is defined as:

\begin{equation}
    \mathcal{L}_{\text{MDM}} = -\mathbb{E}_{t, X_0, X_t} \left[\frac{1}{t} \log p_\theta(X_0 | X_t)\right],
    \label{eq:dlm-obj-ref}
\end{equation}
where $p_\theta(X_0 | X_t)$ is factorized as $\prod_{i=1}^L p_\theta(X_0^i | X_t)$ under standard independence assumptions~\citep{sahoo2024simple}.  
At inference, we begin with a fully masked sequence $X_1$ and iteratively apply the learned reverse process $\log p_\theta(X_0 | X_t)$ to progressively unmask tokens until a clean sequence $X_0$ is obtained.

Most existing MDMs adopt a dense parameterization. At intermediate steps where $0 < t < 1$, all $L$ tokens in $X_t$, both masked and unmasked, are passed to the neural network $p_\theta$, which outputs a dense tensor $y \in \mathbb{R}^{L \times V}$, where $V$ is the vocabulary size. Each $y[i] \in \mathbb{R}^{V}$ represents the logits corresponding to $\log p_\theta(X_0^i | X_t)$.  
This design is computationally inefficient because all $L$ tokens must be processed even when predictions are only needed for a small subset.  
The key contribution of \ours~is a sparse parameterization that directly addresses this inefficiency.

\subsection{Acceleration Methods for MDMs}

MDMs are known to suffer from inefficiencies since they do not natively support KV caching. Several approaches, such as Fast-dLLM~\citep{wu2025fast}, dKV-Cache~\citep{ma2025dkv}, and Sparse-dLLM~\citep{song2025sparse}, attempt to incorporate KV caching into MDMs in a training-free manner through heuristic modifications. However, these methods mostly focus on diffusion large language models (dLLMs) and assume a left-to-right, block-wise, semi-AR sampling scheme. Moreover, while training-free, these approaches often result in unpredictable performance degradation that varies across tasks.

Training-based acceleration methods have also been proposed for MDMs. Prominent examples include Block Diffusion~\citep{arriola2025block}, SDAR~\citep{cheng2025sdar}, and D2F~\citep{wang2025diffusion}, which interpolate between autoregressive and diffusion modeling by introducing block-causal attention masks. Compared with heuristic, training-free methods, these approaches natively support KV caching without inference-time performance degradation and achieve greater speedups by truncating redundant tokens. However, similar to the previous category, they are mostly designed for language modeling and assume a left-to-right semi-AR decoding order. Furthermore, they sacrifice the bidirectional context of MDMs, which is crucial for tasks such as image generation and inpainting.

\ours~is the first method to support both KV caching and token truncation without assuming a left-to-right decoding order or sacrificing bidirectional context. It implements the standard MDM formulation described in \cref{sec:mdm_related} efficiently and faithfully, preserving all desirable properties of MDMs without any qualitycompromise or training–inference gap.

\section{Method}

\subsection{Sparse Parameterization of Masked Sequence}
\begin{figure}
    \centering
    \includegraphics[width=1.0\linewidth]{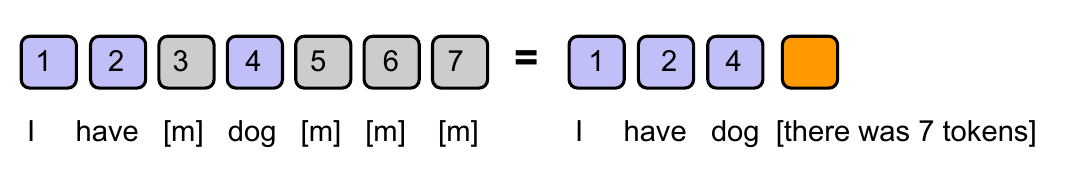}
    \caption{\textbf{Illustration of Sparse Representation of Masked Sequence.} Instead of materializing all masked tokens, a partially masked sequence can be uniquely represented by non-mask tokens, their locations, and the number of total tokens in the original sequence. }
    \label{fig:sparse}
\end{figure}
 The core idea of \ours~is based on the independence of assumption described in \cref{sec:mdm_related}, where $p_\theta(X_0|X_t)$ is factorized into $\prod_{i=1}^L p_\theta(X_0^i|X_t)$. Since $p_\theta(X_0^i|X_t)$ are optimized in the training and sampled at inference independently, if we are only interested in the model's prediction at a subset of locations $C$, there is no reason to predict $p_\theta(X_0^j|X_t)$ where $j\notin C$. However, this observation alone does not allow us to truncate any input tokens, since $X_t^j$ is still part of $X_t$ which are needed to compute $p_\theta(X_0^j|X_t)$.


Recall that $X_t$ is a partially masked sequence containing both clean tokens and mask tokens. Critically, masked tokens carry no substantive information beyond indicating that a position was masked, making compression possible. Consider a seven-token sequence
``I have [m] dog [m] [m] [m]". Rather than representing this with 7 independent tokens, we can equivalently encode it using: (1) clean tokens with their positional embeddings (``I" at position 1, ``have" at position 2, ``dog" at position 4), and (2) a special token indicating the total sequence length (e.g.,~7). The positions of masked tokens are then implicitly determined, as they occupy all positions not taken by clean tokens. In other words, specifying the sequence length is sufficient to represent which positions are masked.

\textbf{Register Tokens.}
In practice, we find that 
using a single special token is not sufficient and leads to considerable performance drop on image generation quality. There are two reasons. First, truncating tokens may leads to considerable drop in model capacity. For example, an 1024$\times$1024 image is represented by 4096 tokens. In the first few sampling steps, we only sample less than 100 tokens. Although aggressively reducing the token count improves efficiency, it does so at the expense of the model’s capacity. Second, previous works\citep{blipdiffusion} using special tokens for text-to-image generation show that a sufficient number of special tokens is needed to meaningfully impact the generation process through attention mechanism. In \ours~, we use 64 special register tokens in our final design, whose position ids are consecutively located at the end of the sequence (in the above example, it will be 8-51). This number remains constant  throughout the inference process and is small in relation to the total sequence length. It does not grow with the number of truncated masked tokens.

\subsection{Sampling with Sparse Parameterization}
\label{sec:sparse-param}
Given a prompt $p$, we sample a sequence of response tokens $X_0$ (image or text) starting from a fully masked sequence $X_1$. We first prefill the KV cache with prompt tokens from $p$. At any sampling step $k$, the model input consists of the prompt $p$, all previously decoded tokens ($C_1 \dots C_{k-1}$), and the tokens to be decoded ($C_k$). Of these inputs, $p$ and $C_1 \dots C_{k-2}$ are already in the KV cache. The \textbf{new cache tokens} $C_{k-1}$ (decoded in the previous step) are processed and added to the cache in this step. Then, the model produces logits only for the \textbf{decode tokens} $C_k$, which are sampled to unmask them. This process repeats until all tokens are unmasked.

To enable proper KV cache updates, we apply a specialized attention mask. Queries from $C_{k-1}$ attend only to \{$p, C_1, \dots, C_{k-1}$\} and cannot attend to $C_k$. This design ensures cached representations are unaffected by masked tokens, enabling efficient training (see \cref{sec:training}). We also include register tokens $R$ at each step, which can attend to all tokens but are only attended to by $C_k$ and $R$. This process is illustrated in Figure \ref{fig:method_infer}.

The remaining 
question 
to be addressed is how to decide the order of unmasking $C_1...C_N$ where $N$ is the number of sampling steps. This strategy differs according to the task.

\textbf{Text-to-Image and Image Editing.} For visual generation tasks, several works \citep{besnier2025halton,li2025lavidao} have proposed to use pre-generated 2D sequences as the unmasking order. 
Compared with confidence-based approaches which dynamically decides which token to unmask at each step based on confidence score, these pre-generated unmasking order have been shown to have higher generation quality. We use the stratified random sampler proposed by LaViDa-O \citep{li2025lavidao}, which generates an unmasking order without the need for confidence scores. This allows us to easily set up $C_1...C_N$ based on a pre-generated unmasking order.
 
\textbf{Text generation and Image understanding.} Unlike image generation and editing tasks, language generation of MDMs typically employ semi-auto-regressive sampling strategy. A sequence of length $L$ is first divided into blocks of equal size $S$. These $\frac{L}{S}$ blocks are sampled auto-regressively from left-to-right. When sampling tokens with in a block, we pass in all tokens in the same block and dynamically decide which tokens to unmask based on confidence scores. In this setup, while we do not know the exact tokens that will be unmasked at $k$-th step $C_k$, we know it belongs to some block $B\supset C_k$. In this step, the input to the model at $k$th step includes the prompt $p$, all previously decoded tokens $C_1...C_{k-1}$ and all tokens in $B$. $C_k\subset B$ is determined after the forward pass based on per-token confidences. While the default left-to-right behavior is similar to Block Diffusion, we highlight that \ours~still supports bi-directional context and arbitrary decoding order of blocks, without assuming a left-to-right block-causal attention masks. This means it can uniquely perform tasks such as text-infilling, constraint generation while methods like block diffusion 
are not capable because they only have one-sided context. Some of these examples are 
shown in \cref{sec:demo}.

\subsection{Training Pipeline}
\label{sec:training}

\begin{figure}[ht!]
    \centering
    \includegraphics[width=0.9\linewidth]{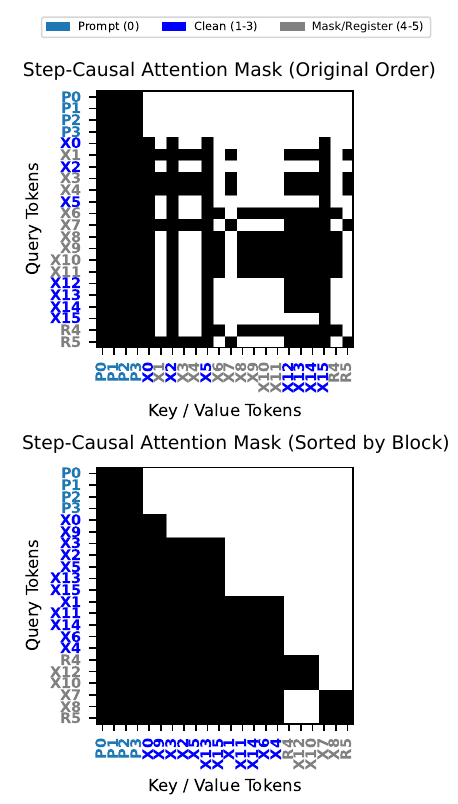}
    \caption{\textbf{Step-Causal Mask.} We employ a step-causal attention mask during training to match the inference behavior of \ours. Consider a sequence containing prompt tokens $P_0$--$P_3$ and answer tokens $X_0$--$X_{15}$, where some tokens are clean and others are masked (color-coded in blue and gray). During inference, prompt tokens and clean tokens are sequentially added to the KV cache.  To simulate this behavior during training, we assign block number~0 to the prompt, and block numbers~1--3 to clean tokens, such that each token may only attend to tokens in its own block or previous blocks.  The bottom figure shows tokens sorted by block number.At inference, the model only observes a subset of masked tokens. To mimic this behavior in training, we assign block numbers~4--5 to masked tokens (e.g., $X_{10}, X_{12}$ in block~4 and $X_7, X_8$ in block~5). Each block is accompanied by a corresponding register token ($R_4, R_5$).  We apply an attention mask such that tokens in one masked block cannot attend to tokens in another masked block, but may attend to all clean and prompt tokens.  This simulates inference paths $0\to1\to2\to3\to4$ and $0\to1\to2\to3\to5$ within a single training step.}
    \label{fig:step_causal_mask}
\end{figure}

      
        

\begin{table*}[t]
\centering
\caption{\textbf{Text to Image Generation Performance on GenEval Dataset}. *These models do not support 1024px generation.}
\label{tab:geneval}
\resizebox{1.0\linewidth}{!}{
\setlength{\tabcolsep}{3pt} 
{\begin{tabular}{lcccccccccc}
\toprule
 & \textbf{Parms} & \textbf{Single}$\uparrow$  & \textbf{Two} $\uparrow$ & \textbf{Position}$\uparrow$  & \textbf{Counting}$\uparrow$  & \textbf{Color}$\uparrow$  & \textbf{Attribution}$\uparrow$  & \textbf{Overall}$\uparrow$  & \textbf{Latency}$\downarrow$  & \textbf{Speedup}$\uparrow$ \\
 \midrule
         SDXL\cite{podell2023sdxl} & 2.6B  & 0.98 & 0.74 & 0.39 & 0.85 & 0.15 & 0.23 & 0.55  & 5.2 & -\\
        DALLE 3\cite{openai_dalle3} & - & 0.96 & 0.87 & 0.47 & 0.83 & 0.43 & 0.45 &  0.67 & - & -\\
        SD3\cite{esser2024scaling-sd3}  & 8B  & 0.99 & 0.94 & 0.72 & 0.89 & 0.33 & 0.60 & 0.74 &23.3 & - \\
        Flux-Dev\cite{flux2024} & 12B & 0.99 & 0.85 & 0.74 & 0.79 & 0.21 & 0.48  & 0.68 & 31.6 & - \\
        Playground v3\cite{li2024playground} & -& 0.99 & 0.95 & 0.72 & 0.82 & 0.50 & 0.54 & 0.76 & - & -\\
                BAGEL \cite{deng2025emerging} & 14B & 0.99& 0.94 &0.64& 0.81 &0.88 &0.63 &0.82 & 45.1 & - \\
                Show-o \cite{xie2024show} & 1B & 0.98 & 0.80 & 0.31 & 0.66 &  0.84 & 0.50 & 0.68 & * & -\\
        MMaDa\cite{yang2025mmada} & 8B & 0.99 & 0.76 & 0.20 & 0.61 & 0.84 & 0.37 & 0.63  & * & - \\

\midrule
LaViDa-O \cite{li2025lavidao} & 10.4B & 0.99  & 0.85 & 0.65 & 0.71 & 0.86 & 0.58 & 0.77 & 21.27 & 1.00 $\times$\\
\rowcolor{gray!20}
\ours  & 10.4B & 0.99 & 0.93 & 0.63 & 0.61 & 0.88 & 0.64 & 0.78 & 10.86 & 1.95 $\times$ \\

\bottomrule
\end{tabular}}}
\end{table*}

\paragraph{Step-Causal Attention Mask.}
To enable efficient parallel training while maintaining consistency with the inference of \ours, we design a step-causal attention mask that simulates the incremental token caching behavior observed during inference.
In a vanilla MDM training step, we have a prompt $p$ and a partially masked response $X_t$, and the model learns to predict the clean sequence $X_0$ given $(p, X_t)$.  
Full attention is applied, and all tokens may interact freely.  
However, this is incompatible with \ours, since KV caching and token truncation imply that during inference:  
(1) not all tokens can attend to each other, and  
(2) the model may not observe all tokens.

To close this training–inference gap, we partition the sequence $X_t$ into $M+N$ blocks and apply a structured attention mask.  
Prompt tokens are assigned block number~0.  
Clean tokens in $X_t$ are randomly assigned block numbers in $\{1,\dots,M\}$.  
Masked tokens are randomly assigned block numbers in $\{M+1,\dots,M+N\}$.

A clean token in block $i \in \{1,\dots,M\}$ may only attend to tokens in blocks $j \le i$.  
For masked tokens in block $i \in \{M+1,\dots,M+N\}$, attention is allowed only to blocks $j=i$ (same masked block) or $j \le M$ (prompt and clean tokens).  
Thus, masked tokens may interact with prompt and clean tokens, but not with masked tokens in other blocks.

For every masked block, we append corresponding register tokens to the sequence and assign them the same block number.  
They follow the same attention rules as masked tokens.  
We note that although we use the term “block,” tokens with the same block number need not form a contiguous segment in $X_t$.  
This design is illustrated in \cref{fig:step_causal_mask}.

\textbf{Training Objective.}
Since \ours~is an alternative parameterization of the standard MDM, it uses the standard MDM training objective.  
At each step, we sample a ground truth sequence $X_0$ from the data distribution (containing both images and text), sample a time $t \in [0,1]$, and draw $X_t \sim q(X_t \mid X_0)$ from the forward masking process.  
We then optimize the MDM objective~\cref{eq:dlm-obj-ref} on the model output $p_\theta(X_0 \mid X_t)$.

The only difference from vanilla MDM training lies in how we implement $p_\theta(X_0^i \mid X_t)$, following our sparse parameterization and step-causal masking.  
Unlike methods such as D2F~\cite{wang2025diffusion}, which rely on distillation-based objectives for post-hoc acceleration, \ours~provides a fundamentally efficient parameterization that supports scalable training and inference without additional distillation stages.

\section{Experiments}

\subsection{Setup}

We initialize \ours~with the pretrained weights of LaViDa-O~\cite{li2025lavidao}, a state-of-the-art 10.4B unified diffusion model that supports a wide range of multimodal tasks, including image understanding, text-to-image generation, and image editing. 

\textbf{Training.} We perform supervised fine-tuning (SFT) on a mixture of image understanding, generation, and editing datasets to adapt LaViDa-O’s dense parameterization to the sparse design of \ours. We source image understanding data from MAmmoth-VL~\citep{guo2024mammoth} and VisualWebInstruct~\citep{visualwebinstruct}. For text-to-image generation, we subsample 20M text–image pairs from LAION-2B~\cite{schuhmann2022laion}, COYO-700M~\cite{kakaobrain2022coyo-700m}, SA-1B~\citep{kirillov2023segment}, JourneyDB~\citep{sun2023journeydb}, BLIP3o-60k~\citep{chen2025blip3}, and ShareGPT4o-Image~\citep{chen2025sharegpt}. We include GPT-Edit-1.5M~\citep{wang2025gpt} for image editing.  
Overall, our training dataset is a filtered subset of the LaViDa-O SFT data, selected for higher quality and efficient fine-tuning. We train for 100k steps on 64 NVIDIA H100 GPUs. Details of our data pipeline and hyperparameters are provided in the Appendix.

\textbf{Evaluation.} We conduct extensive evaluations across diverse multimodal benchmarks to demonstrate the effectiveness of \ours, including GenEval~\cite{ghosh2023geneval}, DPG~\cite{hu2024equipdpg}, and MJHQ-30k~\cite{li2024playground} for text-to-image generation; ImgEdit~\cite{ye2025imgedit} for image editing; and a suite of image understanding benchmarks~\cite{yue2023mmmu,fu2023mme,MMBench,masry-etal-2022-chartqa,mathew2021docvqa,lu2023mathvista,zhang2024mathverse}.  
We also report inference latency (seconds per image) and relative speedup with respect to the base model LaViDa-O. Unless otherwise stated, all image generation experiments are performed at 1024 resolution on a single A100 GPU.

\subsection{Text-to-Image Generation}

We report text-to-image generation results on the GenEval benchmark~\cite{ghosh2023geneval} in Table~\cref{tab:geneval}. We compare against the base model LaViDa-O and other state-of-the-art text-to-image models such as Flux.1-Dev~\cite{flux2024} and unified multimodal models such as MMaDa~\cite{yang2025mmada}.  
\ours~achieves performance comparable to LaViDa-O (+0.01) while substantially reducing end-to-end latency (21.27s vs.\ 10.86s), achieving a $1.95\times$ speedup. It also surpasses models such as Flux.1-Dev in both performance and efficiency, highlighting the effectiveness of our sparse parameterization.

\begin{table}[t]
\centering
\caption{\textbf{Text-to-Image Generation Results on DPG-Bench and MJHQ-30K.} We report the benchmark score of DPG and PickScore, HPS v3, HPS v3, and FID on MJHQ-30k. *We perform SFT on the same data mix as \ours.}
\label{tab:t2i_mjhq}
\resizebox{1.0\linewidth}{!}{
\setlength{\tabcolsep}{1pt}
{\begin{tabular}{lccccc}
\toprule
&  \textbf{DPG}$\uparrow$  & \multicolumn{4}{c}{\textbf{MJHQ-30k}$\uparrow$} \\
\cmidrule(lr){2-2}\cmidrule(lr){3-6}
 & & \textbf{PickScore}$\uparrow$ & \textbf{HPS v2}$\uparrow$  &\textbf{ HPS v3} $\uparrow$ & \textbf{FID}$\downarrow$ \\
\midrule
LaViDa-O \cite{li2025lavidao} & 81.8 & 21.02 & 0.271 & 8.81 & 6.68 \\
LaViDa-O* \cite{li2025lavidao} & 82.1 & 21.04 & 0.297 & 8.87 & 8.11 \\
\rowcolor{gray!20}
\ours & 82.4 & 21.04  & 0.291  &8.89 & 7.63 \\
\bottomrule
\end{tabular}}}
\end{table}
\begin{table*}[h!]
\centering
\caption{\textbf{Image Editing Performance on ImgEdit benchmark.} We report per-category scores and the overall scores.}
\label{tab:image-edit}
\resizebox{1.0\linewidth}{!}{
\setlength{\tabcolsep}{3pt} 
{
\begin{tabular}{lcccccccccccc}
\hline
\textbf{Model} & \textbf{Add} $\uparrow$& \textbf{Adjust} $\uparrow$& \textbf{Extract}$\uparrow$ & \textbf{Replace}$\uparrow$ & \textbf{Remove}$\uparrow$ & \textbf{Background}$\uparrow$ & \textbf{Style}$\uparrow$ & \textbf{Hybrid}$\uparrow$ & \textbf{Action} $\uparrow$& \textbf{Overall}$\uparrow$  & \textbf{Latency}$\downarrow$ & \textbf{Speedup}$\uparrow$ \\
\hline
GPT-4o \citep{openai2024gpt4o} & 4.61 & 4.33 & 2.90 & 4.35 & 3.66 & 4.57 & 4.93 & 3.96 & 4.89 & 4.20 & 111.4 & -\\
Qwen2.5VL+Flux \citep{wang2025gpt} & 4.07 & 3.79 & 2.04 & 4.13 & 3.89 & 3.90 & 4.84 & 3.04 & 4.52 & 3.80 & 55.2 & -\\


FluxKontext dev \citep{labs2025flux1kontextflowmatching} & 3.76 & 3.45 & 2.15 & 3.98 & 2.94 & 3.78 & 4.38 & 2.96 & 4.26 & 3.52& 51.4 & - \\
OmniGen2 \citep{wu2025omnigen2}& 3.57 & 3.06 & 1.77 & 3.74 & 3.20 & 3.57 & 4.81 & 2.52 & 4.68 & 3.44 & 84.8 & -\\
UniWorld-V1 \citep{lin2025uniworld} & 3.82 & 3.64 & 2.27 & 3.47 & 3.24 & 2.99 & 4.21 & 2.96 & 2.74 & 3.26 & 56.2 & -\\
BAGEL \citep{deng2025emerging}& 3.56 & 3.31 & 1.70 & 3.30 & 2.62 & 3.24 & 4.49 & 2.38 & 4.17 & 3.20 & 88.2 & - \\
Step1X-Edit \citep{liu2025step1x}  & 3.88 & 3.14 & 1.76 & 3.40 & 2.41 & 3.16 & 4.63 & 2.64 & 2.52 & 3.06& - & - \\
OmniGen \citep{xiao2025omnigen1} & 3.47 & 3.04 & 1.71 & 2.94 & 2.43 & 3.21 & 4.19 & 2.24 & 3.38 & 2.96 & 126.2 & -\\
UltraEdit \citep{zhao2024ultraedit} & 3.44 & 2.81 & 2.13 & 2.96 & 1.45 & 2.83 & 3.76 & 1.91 & 2.98 & 2.70 & - & -\\
AnyEdit \citep{yu2025anyedit} & 3.18 & 2.95 & 1.88 & 2.47 & 2.23 & 2.24 & 2.85 & 1.56 & 2.65 & 2.45& - & - \\
InstructAny2Pix\citep{li2023instructany2pix} & 2.55 & 1.83 & 2.10 & 2.54 & 1.17 & 2.01 & 3.51 & 1.42 & 1.98 & 2.12 & 48.2 & -\\ 
MagicBrush \citep{zhang2023magicbrush} & 2.84 & 1.58 & 1.51 & 1.97 & 1.58 & 1.75 & 2.38 & 1.62 & 1.22 & 1.90& - & - \\
Instruct-Pix2Pix\citep{brooks2023instructpix2pix} & 2.45 & 1.83 & 1.44 & 2.01 & 1.50 & 1.44 & 3.55 & 1.20 & 1.46 & 1.88 & 9.5 & -\\
\midrule
LaViDa-O \cite{li2025lavidao} & 4.04	&3.62	&2.01&	4.39	&3.98	&4.06	&4.82&	2.94 &	3.54&	3.71 & 63.98 & 1.00$\times$\\
\rowcolor{gray!20}
\ours & 4.08	& 3.73&	2.10	& 4.29&	 3.98	&4.06 &	4.84 &	3.30	&3.76&	3.79 & 22.55 & 2.83$\times$ \\
\hline
\end{tabular}
}
}
\end{table*}

To further assess performance, we evaluate \ours~on DPG-bench~\cite{hu2024equipdpg} and MJHQ-30k~\cite{li2024playground}.  
DPG-bench is evaluated using a VQA model for prompt alignment, while MJHQ-30k reports FID and reward-based metrics including PickScore~\cite{kirstain2023pick}, HPS~v2~\cite{wu2023human}, and the latest VLM-based reward HPS~v3~\cite{ma2025hpsv3} (higher is better).  
As shown in \cref{tab:t2i_mjhq}, \ours~outperforms the LaViDa-O baseline on DPG-bench (+0.6). On MJHQ-30K, \ours~achieves superior results across all perceptual metrics except FID (lower is better), which increases marginally by less than one point.  
Importantly, when both models are trained on the same 20M subset, our Sparse-MDM (FID 7.63) outperforms the baseline LaViDa-O* (FID 8.11), demonstrating that the sparse parameterization and step-causal training not only maintain but can improve generation quality under identical data conditions.

\subsection{Image Editing}

We evaluate image editing performance on the ImgEdit benchmark~\cite{ye2025imgedit}, which measures both visual quality and prompt compliance via a GPT-4 judge model.  
\ours~achieves higher accuracy (+0.08) compared to LaViDa-O and other state-of-the-art unified models such as BAGEL~\cite{deng2025emerging}, while reducing end-to-end latency from 63.98s to 22.55s, achieving a $2.83\times$ speedup.

\subsection{Image Understanding}

To assess text generation efficiency, we evaluate \ours~on the MathVista reasoning benchmark with generation length set to $L=1024$ tokens and block size $S=32$. We compare against LaViDa-O’s vanilla sampling and Fast-dLLM~\cite{wu2025fast}, a training-free KV-caching baseline. Results in Table~\cref{tab:math} show that \ours~matches the base model’s accuracy while achieving a $2.80\times$ speedup.  
It also outperforms Fast-dLLM in both accuracy and latency, confirming the advantage of our learned truncation strategy.

For completeness, we also report \ours~on additional understanding benchmarks including MME-C~\cite{fu2023mme}, MMMU~\cite{yue2023mmmu}, ChartQA~\cite{masry-etal-2022-chartqa}, DocVQA~\cite{mathew2021docvqa}, and MathVerse~\cite{zhang2024mathverse}. These results (Table~\ref{tab:und}) show that \ours~achieves competitive performance across all benchmarks.  
However, speedups are minimal for short QA tasks where outputs contain fewer tokens than one block (32 tokens), effectively reducing \ours~to prompt caching without truncation.

\begin{table}[h!]
\caption{\textbf{Quantative Results on Visual Math Reasoning.} We compare \ours~ with other caching strategies on MathVista accuracy and latency.}
\label{tab:math}
\centering
\resizebox{1.0\linewidth}{!}{
\begin{tabular}{lHHccc}
\toprule
\textbf{Model} & \textbf{Prefix} & \textbf{Suffix} & \textbf{MathVista}$\uparrow$ & \textbf{Latency}$\downarrow$ & \textbf{Speedup}$\uparrow$  \\
\midrule
LaViDa-O \cite{li2025lavidao}            & Cached & --       & 56.9 & 10.41s & 1.00 $\times$\\
LaViDa-O+Fast-dLLM \cite{wu2025fast}   & Cached & Cached    & 56.1 & 5.57s & 1.87 $\times$ \\
\rowcolor{gray!20}
\ours & Cached & Truncated & 56.7 & 3.72s  & 2.80 $\times$\\
\bottomrule
\end{tabular}
}

\end{table}
\begin{table}[h!]
\caption{\textbf{Image Understanding Performance.} We report performance on a wide range of image understanding tasks and compare the performance of \ours with LaViDa-O baseline.}
\label{tab:und}
\centering
\resizebox{1.0\linewidth}{!}{
\setlength{\tabcolsep}{3pt} 
{
\begin{tabular}{lccccccc}
\toprule
\textbf{Model} & \textbf{MME} & \textbf{MMMU} & \textbf{MMB} & \textbf{ChartQA} & \textbf{DocVQA} & \textbf{MathVista} & \textbf{MathVerse} \\
\midrule
LaViDa-O \cite{li2025lavidao} & 488 & 45.1 &76.4 &  80.0 & 73.7 & 56.9 & 36.9 \\
\rowcolor{gray!20}
\ours & 450 & 43.6 & 75.0 & 82.0 & 75.7 & 56.7 & 37.9 \\
\bottomrule
\end{tabular}
}}
\end{table}

\subsection{Qualitative Results}

\begin{figure}[t]
    \centering
    \includegraphics[width=0.9\linewidth]{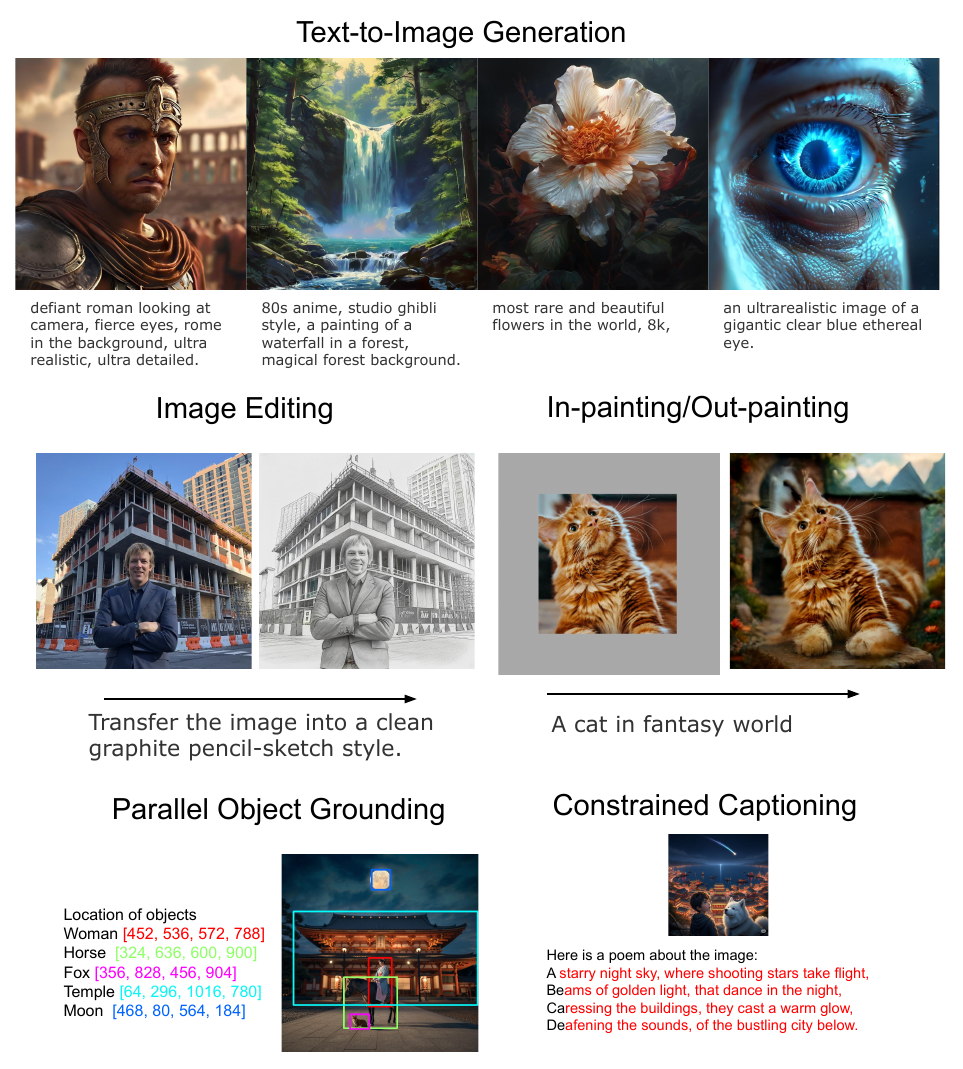}
    \caption{\textbf{Qualitative results.} Unlike semi-AR approaches like Block Diffusion, \ours~supports tasks requiring bidirectional context, such as inpainting/outpainting, parallel grounding, and constrained captioning. In text generation examples, colored regions denote masked tokens initialized for infilling.}
    \label{fig:demo}
\end{figure}

In \cref{fig:demo}, we present qualitative examples across understanding and generation tasks, including text-to-image generation and image editing. Notably, unlike semi-autoregressive methods such as Block Diffusion, \ours~natively supports tasks requiring bidirectional context, such as image inpainting/outpainting, parallel object grounding, and constrained captioning.

\label{sec:demo}

\subsection{Ablation Studies}

\begin{table}[h!]
\centering
\caption{\textbf{Ablation Studies on Speed}. We report the speedup contribution of each key deigns of \ours~on T2I task. }
\resizebox{1.0\linewidth}{!}{
\setlength{\tabcolsep}{3pt} 
{
\begin{tabular}{cccccc}
\hline
\textbf{Cache Prompt} &\textbf{ Cache Res} & \textbf{Truncate Res} & \textbf{Latency} $\downarrow$ & \textbf{Speedup} $\uparrow$ \\
\hline
 &  &  & 21.27 & 1.00 \\
\checkmark &  &  &  16.43 & 1.29 \\
 & \checkmark &  &   18.87 & 1.13 \\
 &  & \checkmark &   17.93 & 1.19 \\
\checkmark & \checkmark &  &  14.09 & 1.51 \\
 & \checkmark & \checkmark &  15.79 & 1.35 \\
\checkmark &  & \checkmark &   13.72 & 1.55 \\
\rowcolor{gray!20}
\checkmark & \checkmark & \checkmark &  10.86 & 1.96 \\
\hline
\end{tabular}
}}
\label{tab:ablation_speed}
\end{table}

\textbf{Effect of token caching and truncation.}  
The speed advantage of \ours~primarily arises from two sources: token caching and truncation. To isolate their contributions, we perform ablations on text-to-image (T2I) tasks. Specifically, we decompose speedup into three components: caching prompt tokens, caching decoded response tokens, and truncating redundant tokens.  
We test all combinations of these components and report results in Table~\cref{tab:ablation_speed}. As shown, enabling any single component improves efficiency, and combining all yields the maximum speedup.

\begin{table}[h]
\centering
\caption{\textbf{Ablation Studies on the Number of Registers.} We report text-to-image generation performance with different number of registers.}
\label{tab:num_reg}
\begin{tabular}{lcccc}
\toprule
\textbf{\#Reg} & \textbf{GenEval} $\uparrow$ & \textbf{DPG} $\uparrow$ & \textbf{HPS v3 }$\uparrow$ & \textbf{FID} $\downarrow$\\
\midrule
0&	0.76&	80.3&	8.68	&9.32\\
1&	0.76&	79.6&	8.71	&9.50\\
32&	0.77&	82.1&	8.87	&8.25\\
\rowcolor{gray!20}
64&	0.78&	82.4&	8.89	&7.63\\
\bottomrule
\end{tabular}
\end{table}

\noindent\textbf{Effect of register tokens.}  
To study the impact of register tokens, we experiment with 0, 1, 32, and 64 registers and report results in Table~\cref{tab:num_reg}.  
We evaluate GenEval and DPG scores as well as FID and HPS~v3 metrics on MJHQ-30k.  
On GenEval, which evaluates high-level prompt alignment via object detection, removing register tokens causes little degradation.  
However, on DPG-bench, which evaluates fine-grained prompt alignment using a VQA model, the absence of register tokens leads to a larger performance drop.  
We also observe measurable differences in image quality metrics (FID and HPS~v3), suggesting that register tokens primarily enhance low-level visual detail rather than high-level structural coherence.

\begin{table}[h]
\centering
\caption{\textbf{Ablation Studies on the Training Strategies.} We demonstrate the effectiveness of our training strategy through performance on GenEval and DPG-bench.}
\label{tab:training_abl}
\begin{tabular}{lcc}
\toprule
\textbf{Model} & \textbf{GenEval} & \textbf{DPG} \\
\midrule
LaViDa-O \cite{li2025lavidao} & 0.77 & 81.8 \\
\midrule
\rowcolor{gray!20}
\ours & 0.78 & 82.4 \\
-No Step Causal Attention Mask & 0.71 & 78.9 \\
-No Training & 0.24 & 47.9 \\
\bottomrule
\end{tabular}
\end{table}

\noindent\textbf{Training strategy.}  
We examine several design choices in our training pipeline, as summarized in Table~\cref{tab:training_abl}. We find that applying the inference pipeline of \ours~to pretrained LaViDa-O without fine-tuning (“No Training”) results in significant performance degradation on GenEval and DPG.  Additionally, removing Step-Causal Attention Mask adversely affect the performance because of mismatched behaviors between training and inference.

\section{Conclusion and Future works.}

In conclusion, we propose \ours, a novel parameterization for multi-modal MDMs. It offers significant speedup on a wide range of visual and text generation tasks such as text-to-image generation, image editing, and visual math reasoning without compromising generation quality.  Despite promising results, \ours~has several limitations. First, while \ours~ offers a significant speedup, it requires additional training. We emphasize that \ours~offers faster speedup than most aggressive KV caching strategy that caches all possible tokens (\cref{tab:ablation_speed}), which is an upper-bound for all heuristic-based training-free methods.  Second, while in principle \ours~is just an efficient parameterization for the standard MDM and can be used to pre-train large models from scratch, we conducted our experiments in a post-trianing setup due to compute costs. In future we will explore additional scaling and train from scratch.

\section{Acknowledgement}

AG would like to acknowledge the support from Schmidt Sciences and NSF Career Award \#2341040. SL is in part supported by Amazon Fellowship.


{
    \small
    \bibliographystyle{ieeenat_fullname}
    \bibliography{main}
}
\clearpage
\setcounter{page}{1}
\onecolumn
   {
        \centering
        \Large
        \textbf{\thetitle}\\
        \vspace{0.5em}Supplementary Material \\
        \vspace{1.0em}
   }   

\newcommand{\cat}[0]{\text{Cat}}
\newcommand{\alphats}[0]{\frac{1-t}{1-s}}
\newcommand{\oneminusalphats}[0]{\frac{t-s}{1-s}}

\section{Additional Technical Details}

\subsection{Formulation of Discrete Diffusion Models}

In this section, we include an overview of the standard formulation of Masked Diffusion Models (MDMs) that are widely adopted by literature \cite{sahoo2024simple,lou2023discrete-sedd,you2025lladav,li2025lavida,li2025lavidao}. 

Given a  sequence $X_0$ consisting of discrete tokens $[X_0^1,X_0^2,\ldots,X_0^L]$, where $L$ is the sequence length, the forward process $q(X_t|X_s)$ gradually mask the sequence and convert clean tokens to a special mask token $[M]$ over the continuous time interval $[0,1]$, with $1 \ge t \ge s \ge 0$. At $t=1$, the sequence $X_1=[X_1^1,X_1^2,\ldots,X_1^L]$ consists of only masked token. This forward process is formally defined as


\begin{equation}
    q(X_t^i|X_s^i) =  
    \begin{cases}
      \cat(X_t^i;\textbf{M}), & \text{if } X_s^i=[M] \\
      \cat(X_t^i;\alphats \mathbf{X_s^i}+\oneminusalphats \textbf{M}), & \text{if } X_s^i \ne [M],
    \end{cases}
\end{equation}

where $\cat(\cdot)$ denotes a categorical distribution, and $\textbf{M}, \mathbf{X_0^i}, \mathbf{X_s^i} \in \mathbb{R}^{|V|}$ are probability vectors, and $|V|$ is the vocabulary size. In particular, $\textbf{M}$ is a one-hot vector corresponding to the special token $[M]$. This forward process yields the following marginal distribution:

\begin{equation}
    q(X_t^i|X_0^i) =  \cat(X_t^i;(1-t) \mathbf{X_0^i}+t \textbf{M}).
    \label{eq:q_process}
\end{equation}

Prior works \citep{sahoo2024simple} show that the posterior of the reverse process $p(X_s|X_t,X_0)$ has the following form:

\begin{equation}
    p(X_s^i|X_t^i,X_0^i) =  
    \begin{cases}
      \cat(X_s^i;\mathbf{X_t^i}), & \text{if } X_s^i \ne [M] \\
      \cat(X_s^i;\tfrac{t-s}{t} \mathbf{X_0^i}+\tfrac{s}{t} \textbf{M}), & \text{if } X_s^i = [M].
    \end{cases}
    \label{eq:appendix-eq-p}
\end{equation}

At inference, $\mathbf{X_0}$ is not known, so we replace $\mathbf{X_0^i}$ with the neural network prediction $p_\theta(X_0^i|X_t)$, which gives the following emprical sampling process:

\begin{equation}
    p_\theta(X_s^i|X_t) =  
    \begin{cases}
      \cat(X_s^i;\mathbf{X_t^i}), & \text{if } X_s^i \ne [M] \\
      \cat(X_s^i;\tfrac{t-s}{t} p_\theta(X_0^i|X_t)+\tfrac{s}{t} \textbf{M}), & \text{if } X_s^i = [M].
    \end{cases}
    \label{eq:appendix-inference}
\end{equation}

\textbf{Sampling process.} To sample a sequence of clean tokens $X_0$, we start with a fully masked sequence $X_1$,  where $X_1^1=\cdots=X_1^L=[M]$. We discretize the continuous time interval $[0,1]$ into discrete timesteps $0=t_0<t_1<\cdots<t_K=1$, and iteratively sample $X_{t_{k-1}}\sim p_\theta(X_{t_{k-1}}|X_{t_k})$ using Equation \ref{eq:appendix-inference}. We start with $k=K$ and end when we obtain a mask-free sequence $X_0$. At each step, we assume $p_\theta(X_{t_{k-1}}|X_{t_k})$ factorizes as $\prod_{i=1}^L p_\theta(X_{t_{k-1}}^i|X_{t_k})$, following previous works \citep{nie2025large, sahoo2024simple, lou2023discrete-sedd}. 

\textbf{Training process.} At each training step, given a clean sequence $X_0$, we sample a random timestep $t\in[0,1]$ and obtain $X_t\sim q(X_t|X_0)$ through the forward process defined in Equation \ref{eq:q_process}. This gives us a partially masked sequence. The loss is then computed using Equation \ref{eq:dlm-obj-ref} from Section \ref{sec:mdm_related}.

We highlight that \ours~does not fundamentally change these formulations. Rather, it proposes an equivalent but efficient parameterization by introducing a sparse representation for the partially masked sequence $X_t$.

\clearpage

\begin{figure*}[h]
    \centering
    \includegraphics[width=1.0\linewidth]{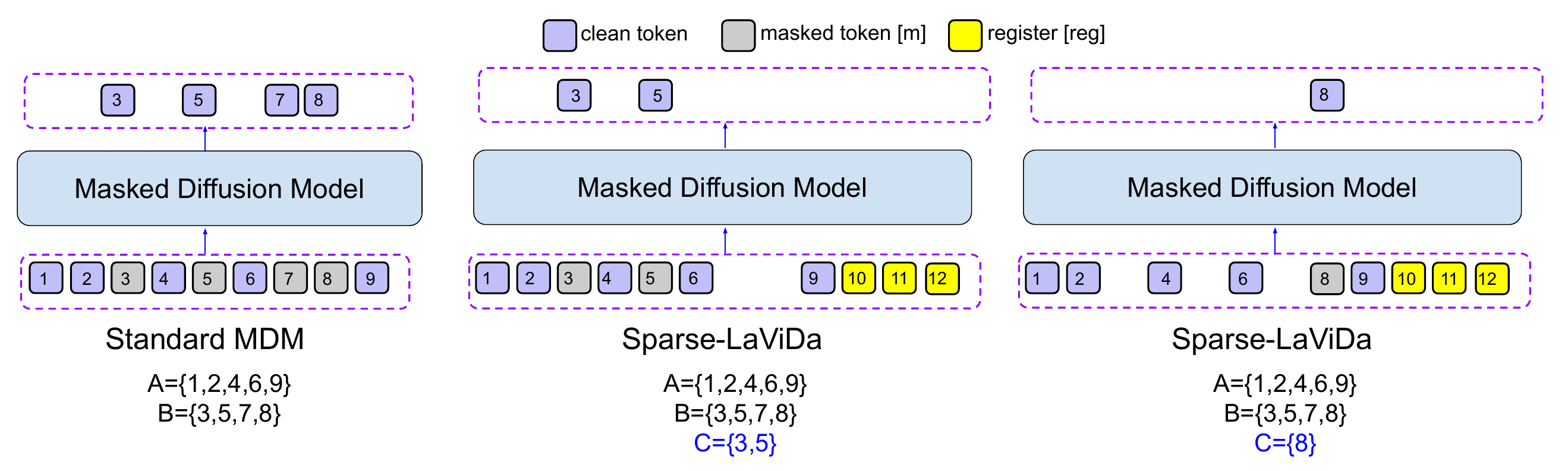}
    \caption{\textbf{Sparse-Representation of Partially Masked Sequence.} Given a partially masked sequence, we can partition its tokens into two subsets $A$ and $B$, where $A$ consists of clean tokens and $B$ consists of masked tokens. The standard MDM (Left) need to pass all $|A|+|B|$ tokens. By contrast, Sparse-LaViDa (Middle and Right) only need to pass $|A|+|C|+m$ tokens, where $C$ is a subset of $B$ and $m$ is the number of register tokens (set to 3) in this case.   }
    \label{fig:appendix-token}
\end{figure*}
\subsection{Register Tokens}

In this section, we discuss the detail implementation of register tokens. Given a partially masked sequence $X_t$ of length $L$, which can be partitioned into two subsets $X^A,X^B$, where $A$ consists of clean token positions and $B$ consists of masked positions. Let $K=\{1,2..L\}$ be the set of all token positions, we have that $A\cup B=K$ and $A\cap B=\emptyset $. In the standard parameterization, we need to pass in $L=|A|+|B|$ tokens, among which $|B|$ are masked tokens, to the model, even if we are only interested in obtaining the prediction at a subset of masked position $C\subset B$ at the current diffusion step. 

The sparse parameterization introduces an extra set of $m$ register tokens $R=[R^1..R^m]$. Concretely, these are achieved by adding special tokens ``[reg]" that is similar to mask token ``[M]" to the vocabulary. All $m$ register tokens are represented with the same special token ``[reg]"  at the tokenization level, but their positional ids are $L+1,L+2...L+m$ respectively, leading to different rope embeddings and different beahvior in the attention process. In this setup, the total token count is $|A|+|C|+m$. When $|C| \ll |B|$, we can achieve considerable speedup from reduced sequence length. 

We visualize this design in \cref{fig:appendix-token}. Sparse-LaViDa allows us to flexibly truncate masked tokens depending on which token prediction we want to obtain.  When, $C=B$, Sparse-LaViDa reduces to the standard MDM parameterization. Hence, Sparse-LaViDa is a generalization to the standard MDM.

For simplicity, we use a simple partition $A$, $B$ to elaborate the design of register tokens and sparse representation, as opposed to more complex partition $C_1 . . . C_{k-1}$ that depends on sampling steps, which are used in \cref{sec:sparse-param} of the main text and the following \cref{sec:attn-mask-appendix} of the appendix. 

\subsection{Step-Causal Attention Masks}

\begin{figure*}[ht]
  \centering
  \begin{subcaptionbox}{\textbf{Inference with Sparse Parameterization} \label{fig:appendix-mask-inf-a}}[0.64\textwidth]
    {\includegraphics[width=\linewidth]{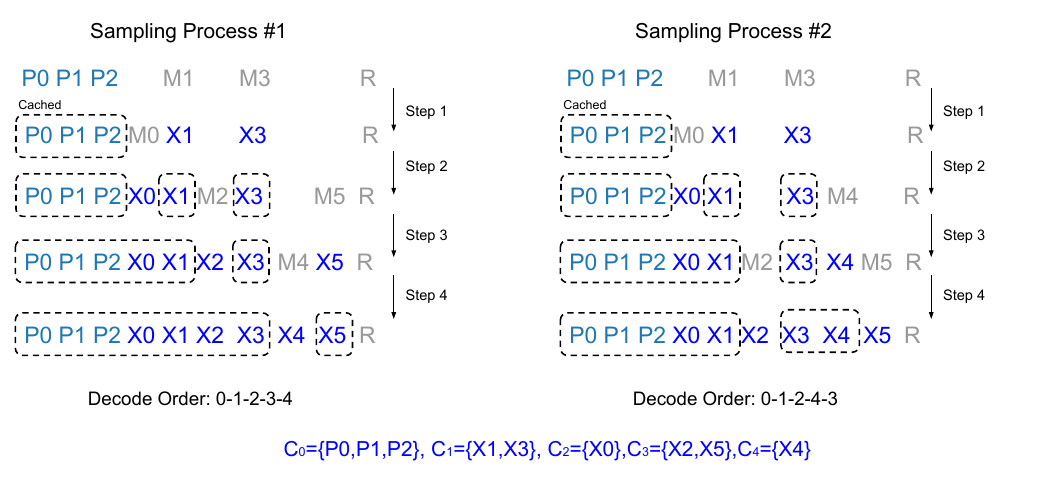}}
  \end{subcaptionbox}
  \begin{subcaptionbox}{\textbf{Design of Step-Causal Attention Mask} \label{fig:attn-mask-b-appendix}}[0.32\textwidth]
    {\includegraphics[width=\linewidth]{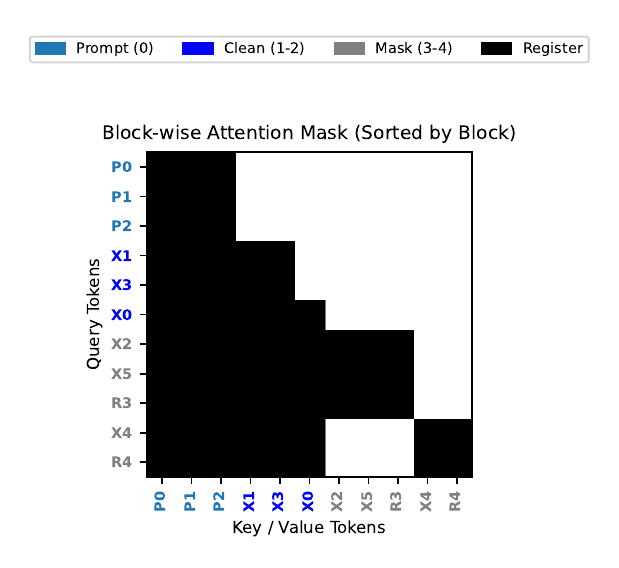}}
  \end{subcaptionbox}
  \caption{\textbf{Connection between the Diffusion Sampling Process and the Step-Causal-Attention-Mask.} (Left) Given prompt tokens $P_0...P_2$ and response tokens $X_0...X_5$, we partition the response tokens with $C_1...C_4$ and assign prompt tokens to $C_0$. Both 0-1-2-3-4 and 0-1-2-4-3 are viable sampling order of tokens in a 4-step diffusion process. We use M to represent mask tokens and R represent registers. (Right), for the two sampling process, we can simulate step 3 of both processes simultaneously via a special attention mask.  }
  \label{fig:method_infer_appendix}
\end{figure*}

In this section, we provide a detailed account of step-causal attention mask, with particular focus on how the design of  step-causal attention mask aligns with the inference process with KV caching. 

\textbf{Inference Time.} Suppose we have $S$ prompt tokens and generate a clean response with $L$ tokens through $k$ diffusion steps. We can naturally obtain a $k$-partition of the $L$ response tokens $C_1..C_k$ depending on when they are unmasked during the $k$ diffusion steps. We also define $C_0$ as the set of prompt tokens, which are never masked. 

\cref{fig:appendix-mask-inf-a} illustrates an example, which consists of $S=3$ prompt tokens $P_0...P_2$ and $L=6$ response tokens $X_0...X_5$.  We also assume we have $k=4$ diffusion steps and there is only one register token $R$. As illustrated in the figure, sampling process $\#1$ induces the partition $C_0=\{P_0,P_1,P_2\}$,  $C_1=\{X_1,X_3\}$, $C_2=\{X_0\}$, $C_3=\{X_2,X_5\}$, $C_4=\{X_4\}$, where tokens in $C_1,C_2,C_3,C_4$ are unmasked at the first, second, third, and fourth diffusion steps respectively. After a token is generated at $i$th step, it is passed back to the model in $(i+1)$th step and added to the cache after the model forward call. This is also illustrated in the figure (black dashed box). For example, tokens in $C_1=\{X_1,X_3\}$ is added to the KV cache after step 2.

At the $i$th diffusion step, the KV cache consist of all tokens in $C_j$ where $j < i-1$.  The input of the model consist of the previously generated tokens in $C_{i-1}$, the set of masked tokens to be decoded at the current step $C_i$, and  the register token $R$. Since $C_{i-1}$ are to be added to the KV cache, we prevent the queries of $C_{i-1}$ to interact with keys and values of $C_i$ and $R$ as discussed in \cref{sec:training}. 

Taking step 3 as an example, the KV cache consists of tokens in $C_0,C_1$ (boxed), which are $P_0,P_1,P_2,X_1,X_3$. The input consists of tokens in $C_2$, which is just $X_0$, and masked tokens in $C_3$, denoted by $M_2,M_5$, as well as a register token $R$. In this step, $X_0$ from $C_2$ cannot attention to $M_2,M_5,R$, (color-coded as gray), while  $M_2,M_5,R$ can attend to all other tokens. 

Note that while the partition $C_0,C_1...C_4$ derives from the sampling process $\#1$, its decoding order $0-1-2-3-4$ does not uniquely leads to this partition. For instance, sampling process $\#2$ with decoding order $0-1-2-4-3$ also has the same partition  (\cref{fig:appendix-mask-inf-a}, Right).  Moreover, we note that at step 3 of  sampling process $\#2$, the KV cache is exactly the same as sampling process $\#1$. Among the input tokens, $X_0$ from $C_2$ has exactly the same behavior as sampling process $\#1$ because of the attention mask. The only differences lies in the input of masked tokens. In sampling process $\#1$, we have $M_2,M_5$ since at this step we want to unmask $C_3$. In sampling process $\#2$, we have $M_4$ since at this step we want to unmask $C_4$. This observation gives us the opportunity to parallelize the training by simulate sampling process $\#1$ and sampling process $\#2$ simultaneously through a special attention mask.

\textbf{Training time.} We can merge sampling process $\#1$ and sampling process $\#2$ by constructing the sequence $P_0,P_1,P_2,X_1,X_3,X_0,X_2,X_5,R_3,X_5,R_4$, where $R_3,R_4$ are duplicates of the same register token $R$. We assign a step-causal attention mask for tokens $P_0,P_1,P_2,X_1,X_3,X_0$, as they have exactly the same attention pattern in both sampling steps.  To accommodate the difference of masked token inputs where sampling process $\#1$ has $M_2,M_5,R$ and sampling process $\#2$ has $M_4,R$ we concatenate the sequences to form $M_2,M_5,R_3,M_4,R_4$ and design a block-diagonal attention mask that ensures input tokens from one sampling process cannot attend to tokens from another same sampling process  (bottom-right section of \cref{fig:attn-mask-b-appendix}). We denote two copies of $R$ as $R_3,R_4$ since $M_2,M_5$ comes from partition $C_3$ and  $M_4$ comes from partition $C_4$. This design of attention mask is illustrated in \cref{fig:attn-mask-b-appendix}.

The training-time attention mask is formally defined in \cref{alg:step_causal_mask}. In the above example, to simulate sampling process $0-1-2-3-4$ and $0-1-2-4-3$, we set the number of clean blocks $M=2$ (i.e. $C_0,C_1$) and the number of masked blocks $N=2$  (i.e. $C_3,C_4$). We assign the block number to tokens based on which of the partition $C_0...C_4$  they belong to. The duplicated register tokens $R_3,R_4$ are assigned with block numbers 3 and 4 respectively.

\begin{algorithm}[H]
\caption{Step-Causal Attention Mask Construction}
\label{alg:step_causal_mask}
\begin{algorithmic}[1]
\Require Block assignments $A \in \mathbb{Z}^{L+S}$, number of clean blocks $M$, number of masked blocks $N$
\State Initialize $\text{mask} \in \{0,1\}^{(L+S) \times (L+S)}$ to zeros

\For{$q_i = 0$ \textbf{to} $L+S-1$} 
    \State $q_b \gets A[q_i]$ \Comment{Block index of query token}

    \For{$k_j = 0$ \textbf{to} $L+S-1$}
        \State $k_b \gets A[k_j]$ \Comment{Block index of key token}

        \If{$q_b = 0$} \Comment{Prompt token: full attention to prompt}
            \If{$k_b \le 0$}
                \State $\text{mask}[q_i, k_j] \gets 1$
            \EndIf

        \ElsIf{$1 \le q_b \le M$} \Comment{Clean token: attend up to its block}
            \If{$k_b \le q_b$}
                \State $\text{mask}[q_i, k_j] \gets 1$
            \EndIf

        \ElsIf{$M+1 \le q_b \le M+N$} \Comment{Masked tokens: attend to prompt/clean tokens, or those in the same block}
            \If{$(k_b \le M) \;\lor\; (k_b = q_b)$}
                \State $\text{mask}[q_i, k_j] \gets 1$
            \EndIf
        \EndIf

    \EndFor
\EndFor

\State \Return mask
\end{algorithmic}
\end{algorithm}

\label{sec:attn-mask-appendix}
\section{Additional Experiment Details and Results}

\subsection{Data pipeline}

Our training dataset consist of the following tasks.
\begin{itemize}
    \item \textit{A: Text-to-Image Pairs.} We source data from LAION-2B \citep{schuhmann2022laion} and COYO-700M \citep{kakaobrain2022coyo-700m}. We additionally include BLIP3o-60k \citep{chen2025blip3}, and ShareGPT4o-Image \citep{chen2025sharegpt}. Each dataset is heavily filtered to remove NSFW prompts, low CLIP scores \citep{radford2021learning}, low aesthetic scores \citep{laion-aesthetics}, and low-resolution images following LaViDa-O's pipeline. We include all data from BLIP3o-60k and ShareGPT4o-Image, and select highest-quality samples from LAION and COYO based on CLIP scores and aesthetic scores.  The final data mix consist of 20M images. Unlike LaViDa-O, we did not include  SA-1B \citep{kirillov2023segment}, JourneyDB \citep{sun2023journeydb} because of quality issues. Specifically, human faces in SA-1B are blurred, while JourneyDB images have many artifacts even after heavy filtering. 
    
    \item \textit{B: Image-level Understanding Data.} We include MAmmoth-VL \citep{guo2024mammoth}, and VisualWebInstruct \citep{visualwebinstruct}. 
    \item \textit{C: Region-level Understanding Data.} We include GranD \citep{hanoona2023GLaMM} and RefCOCO \citep{kazemzadeh2014referitgame}.
    \item \textit{D: Image Editing Data.} We include ShareGPT4o-Image \citep{chen2025sharegpt}, GPT-Edit-1.5M \citep{wang2025gpt}, and the image editing subset of UniWorld-V1 \citep{hu2022unified}.
 \end{itemize}

\subsection{Training Setup}

We perform SFT training on the LaViDa-O weights using our proposed sparse-parameterization and step-causal attention mask. The hyper-parameters largely followed the SFT stage of LaViDa-O. We document the details of training setup and relevant hyperparameters in \cref{tab:training-stages}. The main experiments is conducted with 64 A100 GPUs across 8 nodes. The training takes 5 days in total, which is 15\% of the LaViDa-O's training budget (34.2 days from scratch).

\begin{table}[h]
\centering
\caption{\textbf{Training configurations of \ours.} We report the relevant hyperparameters for training, including the learning rate, number of training steps, optimizer setup, image resolution for understanding and generation tasks. }
\label{tab:training-stages}
\begin{tabular}{lHHc}
\toprule

 & \textbf{Stage 1} & \textbf{Stage 2} & \textbf{SFT} \\
\midrule
Learning Rate & $5 \times 10^{-6}$ & $1 \times 10^{-4}$ & $2 \times 10^{-5}$ \\
Steps & 80k & 400k & 100k \\
$\beta_1$ & 0.99 &0.99  & 0.99 \\
$\beta_2$ & 0.999 &0.999  & 0.999 \\
optimizer & AdamW & AdamW & AdamW \\
\midrule
Loaded Parameters & 8B & 6.4B & 10.4B \\
Trainable Parameters & 8B & 2.4B & 10.4B \\
Und. resolution & 384 $\times \{(1,3),(2,2)\}$ & 384 $\times \{(1,3),(2,2)\}$ & 384 $\times \{(1,3),(2,2)\}$\\
Gen. resolution & - & 256 $\rightarrow$ 512 $\rightarrow$ 1024 & 1024 \\
\midrule
\end{tabular}%

\end{table}

\subsection{Additional Results on Object Grounding}

As shown in \cref{fig:demo}, one of the advantages of \ours~is that it does not employ a block-causal attention mask like block-diffusion, making it capable of tasks that requires bi-directional context such as inpainting, object grounding through infilling, and constrained text generation just like vanilla MDM. 

We report quantitative results on object grounding tasks in Table \ref{tab:grounding}. Overall, \ours~achieves comparable performance as the LaViDa-O baseline. Since all coordinates are generated in a single step, the two methods have identical inference speed because \ours~cannot benefit from token truncation in this setup.

\begin{table}[h!]
\centering
\caption{\textbf{Precision@0.5  on RefCOCO, RefCOCO+, and RefCOCOg REC tasks.} }
\label{tab:grounding}
\setlength{\tabcolsep}{10pt} 
{
\begin{tabular}{lccccccccH}
\hline
\textbf{Model} & \multicolumn{3}{c}{\textbf{RefCOCO}}$\uparrow$ & \multicolumn{3}{c}{\textbf{RefCOCO+}}$\uparrow$ & \multicolumn{2}{c}{\textbf{RefCOCOg}}$\uparrow$  & \textbf{Latency}$\downarrow$ \\
\cmidrule(lr){2-4} \cmidrule(lr){5-7} \cmidrule(lr){8-9}
 & val & testA & testB & val & testA & testB & val & test & (s/image) \\
\hline
SegLLM-7B\citep{wang2025segllm} & 90.0 & 92.1 &  86.2 & 82.2& 85.5 &76.1 & 83.9 & 85.9 \\
Qwen2.5-VL-7B \citep{bai2025qwen25-vl} & 90.0 & 92.5 & 85.4 & 84.2 & 89.1 & 76.9 & 87.2 & 87.2 \\
GroundingDINO \citep{liu2024grounding}& 90.6 & 93.2 & 88.2 & 88.2 & 89.0 & 75.9 & 86.1 & 87.0 \\
InternVL3-8B \citep{zhu2025internvl3} & 92.5 & 94.6  & 88.0 & 88.2 & 92.5 & 81.8 & 89.6 & 90.0 \\
\midrule

LaViDa-O & 91.9&	94.6	&88.4	&87.4 &	91.7 &	82.2	 &89.5	&89.8 \\
\rowcolor{gray!20}
\ours & 92.3&	94.9	&89.1	&87.4 &	92.5 &	81.8	 &89.6	&89.9 \\
\hline
\end{tabular}
}
\end{table}

\subsection{Addtional Results with Larger Batch Size}

In Table \ref{tab:throughput}, we report results at a larger batch size for T2I tasks. Results show that LaViDa-O consistently outperforms baselines at varying batch sizes. 

\begin{table}[h]
    \centering
    \begin{tabular}{ccc}
    \toprule
        \textbf{Bs}. & \textbf{LaViDa-O } $\uparrow$  &\textbf{ \ours} $\uparrow$    \\
        \midrule
        1 & 2.82 & 5.52 \\
        2 &  3.85  & 7.26  \\
        4 &  4.24  & 8.65  \\
        \hline
    \end{tabular}
    \caption{\textbf{Throughput Analysis. We report images/min. (Higher is better)}}
    \label{tab:throughput}
\end{table}

\subsection{Additional Qualitative Comparisons}

\begin{figure}
    \centering
    \includegraphics[width=1.0\linewidth]{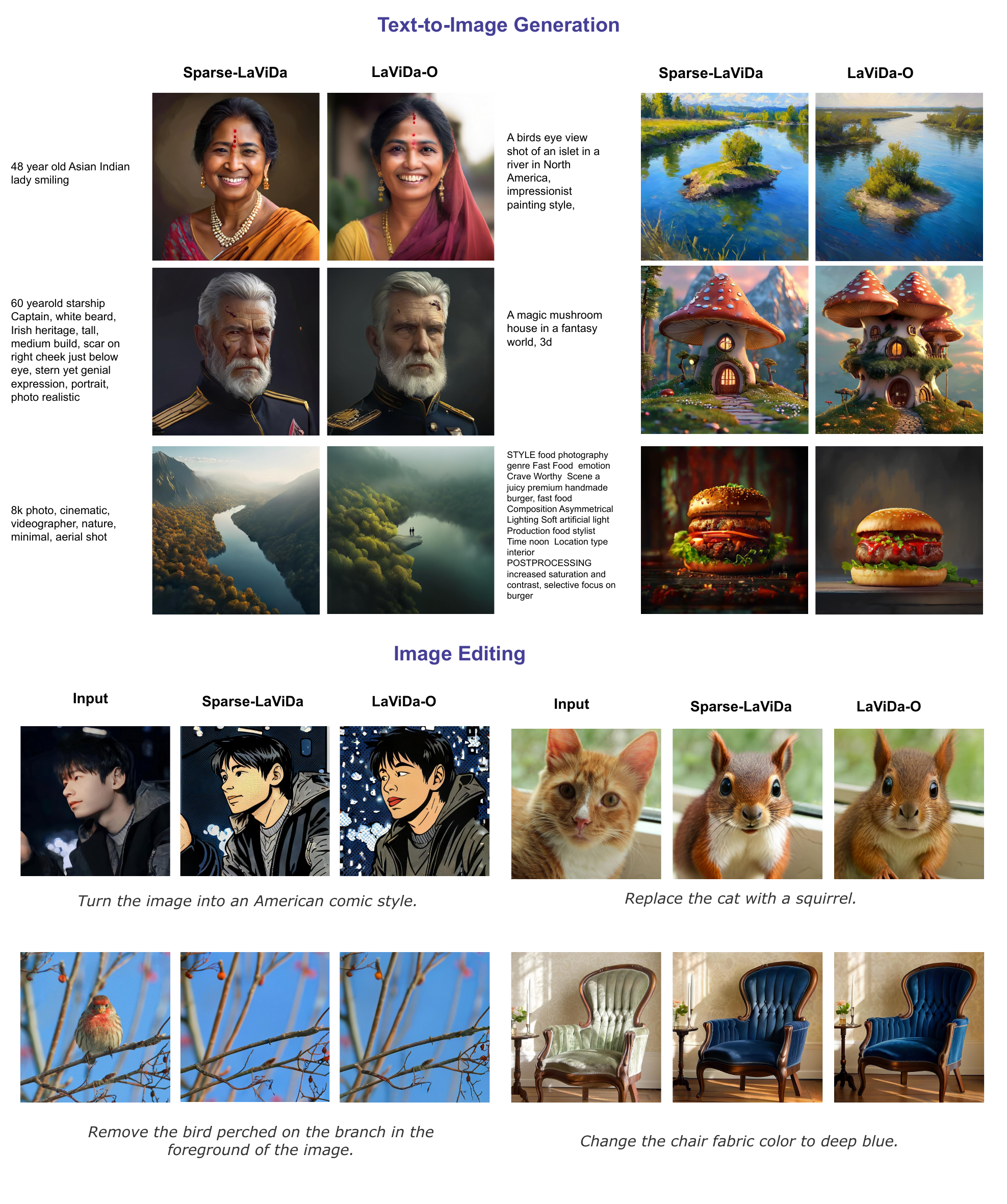}
    \caption{\textbf{Qualitative Comparisons of \ours~and LaViDa-O baseline.} We show qualitative results of text-to-image generation and image editing results of \ours~and LaViDa-O baseline.\ours~achieves a speedup of $1.95\times$ on text-to-image generation and a speedup of $2.83\times$ on image editing, while maintaining comparable visual quality   }
    \label{fig:quality-appendix}
\end{figure}

In Figure \ref{fig:quality-appendix}, we provide side-by-side qualitative comparisons of \ours~and the LaViDa-O baseline on text-to-image generation and image editing tasks. As shown in the figure, \ours~ achieves comparable generation quality while offering  a speedup of $1.95\times$ on text-to-image generation and a speedup of $2.83\times$ on image editing.

\section{Limitations}

Despite the promising results of \ours, it has several limitations. First, the speedup only benefits long sequence generation such as text-to-image generation, image-editing, or visual math problem-solving with long reasoning chains. It does not improves the speed of short QA tasks or object grounding tasks, where the number of output tokens are very small. We emphasize that \ours~will benefit any tasks involving image generation, since an image is represented by 4096 VQ tokens.

Second, our model inherits many limitations from the base model LaViDa-O, such as hallucination in responses. Further, we observe the same pixel shift issue discovered by LaViDa-O, where image editing results may have subtle changes in un-edited regions.  We hope this issue will be addressed by future foundational models.

Lastly, while we have demonstrated \ours~is an efficient post-training technique, in principle it should be general to all stages including pretraining. It would be interesting for researchers with more compute resources to apply this to different stages of training. 

\end{document}